%% file: IEE_Ijcnn.tex
\documentclass[conference]{IEEEtran}
\makeatletter
\newcommand{\linebreakand}{%
  \end{@IEEEauthorhalign}
  \hfill\mbox{}\par
  \mbox{}\hfill\begin{@IEEEauthorhalign}
}
\makeatother

\IEEEoverridecommandlockouts
\usepackage{cite}
\usepackage{amsmath,amssymb,amsfonts}
\usepackage{algorithmic}
\usepackage{graphicx}
\usepackage{textcomp}
\usepackage{xcolor}
\def\BibTeX{{\rm B\kern-.05em{\sc i\kern-.025em b}\kern-.08em
    T\kern-.1667em\lower.7ex\hbox{E}\kern-.125emX}}
\begin{document}

\title{Evaluating Spiking MoveNet Deployability for Near-Future Onboard Spiking Satellite\\
Pose Estimation
\thanks{This material is based upon work supported by the French national research agency (project DEEPSEE ANR-17-CE24-0036) and by the donation from Franck DIARD to the Universite Cote d'Azur.
This work is carried out as part of a collaboration between LEAT and IRT Saint Exupéry.}
}

\author{
    \IEEEauthorblockN{Jonathan COURTOIS}
    \IEEEauthorblockA{\textit{Université Côte d'Azur$^1$, CNES} \\
    Biot, France}
    \and
    \IEEEauthorblockN{Christophe MARABOTTO}
    \IEEEauthorblockA{\textit{IRT Saint Exupéry$^2$} \\
    Biot, France}
    \and
    \IEEEauthorblockN{Benoit MIRAMOND}
    \IEEEauthorblockA{\textit{Université Côte d'Azur$^1$} \\
    Biot, France}
    \and
    \IEEEauthorblockN{Alain PEGATOQUET}
    \IEEEauthorblockA{\textit{Université Côte d'Azur$^1$} \\
    Biot, France}
    \linebreakand
    \IEEEauthorblockA{\textit{$^1$name.surname@univ-cotedazur.fr} \\
    \textit{$^2$name.surname@irt-saintexupery.com}}
}


\maketitle

\input{sec2/0_abstract}


\input{sec2/1_intro}
\input{sec2/2_related_work}

\input{sec2/3_experiments}
\input{sec2/4_conclusion}

\end{document}

%% file: sec2/0_abstract.tex
\begin{abstract}
SNNs are attracting interest due to their potential for energy efficiency and event-driven processing. However, neuromorphic hardware is not yet mature enough to enable embedded deployment of complex spiking network architectures.
In this work, we investigate a spiking version of the MoveNet architecture tailored for autonomous spacecraft operation, aiming to evaluate the deployability of this complex architecture compared to the non-spiking version for embedded, fully event-based solutions in space applications.
Using the RAPTOR dataset, we compare the performance and operational cost of the proposed S-MoveNet-Lightning architecture with its standard version, MoveNet-Lightning.
Our results show balanced advantages in computational resources and architectural settings for spiking U-Net-like architectures.
An evaluation of the performance and hardware footprint of the proposed Spiking Neural Network (SNN) approach for pose estimation shows that, on average, 2.2 times fewer multiplication and accumulation operations are required compared to the non-spiking solution deployed on FPGA platforms with the AMD Deep Learning Processor Unit (DPU).
These results provide valuable insights into the feasibility and potential of SNNs in complex aerospace applications and highlight their role in balancing efficiency and performance, paving the way for event-driven solutions in embedded space applications.

\end{abstract}

%% file: sec2/1_Intro.tex
\section{Introduction}
\label{sec:intro}

\begin{figure}[ht]
    \vspace{0.3cm}
    \centering
    \includegraphics[width=0.95 \linewidth]{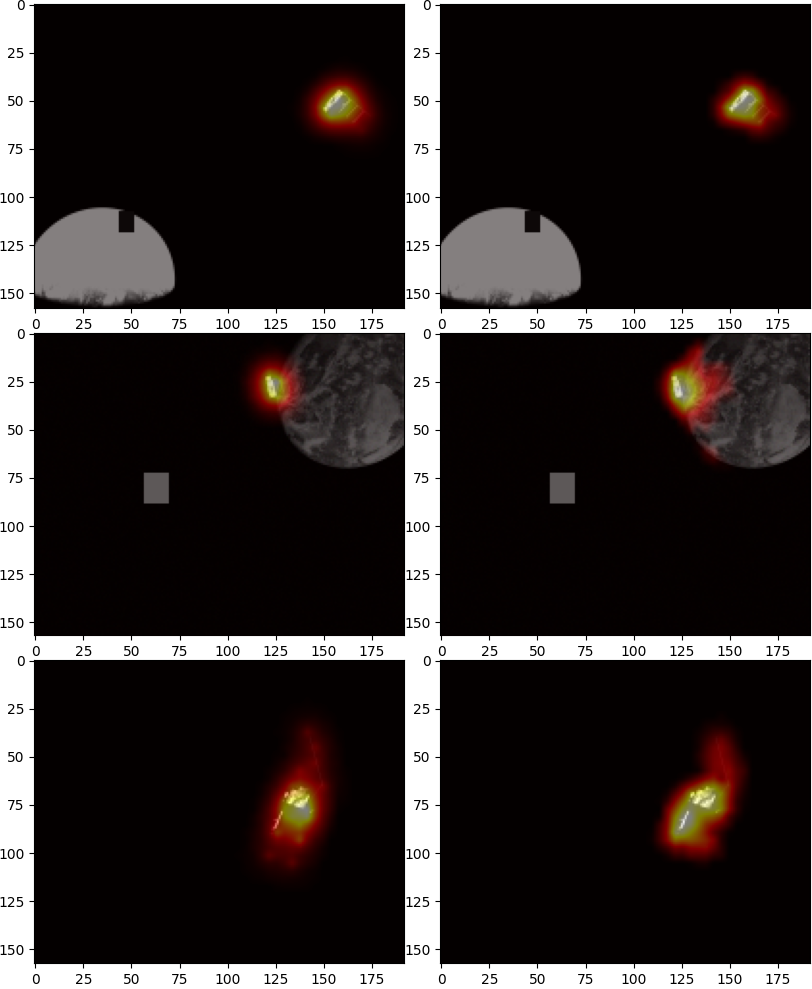}
    \caption{Stacked heatmaps visualization in comparison between the ground truth (left) and the Spiking MoveNet-Lightning output (right). The visualization samples were extracted from the training set with data augmentation (Top 2) and validation set (bottom).}
    \label{fig:Heatmapout}
\end{figure}

Targeting efficient on-board event-based solutions assisted by neural networks requires three key components: event-based datasets, event-based neural architectures, and neuromorphic hardware. Although the space industry has shown increasing interest in event-based applications, such as Earth-based Space Situational Awareness (SSA) systems \cite{afshar2020event}, Earth Observation (EO) with the first satellite on-board SNN \cite{abderrahmane2022spleat}, event-based sensing applications \cite{mcharg2022falcon}, and event-based pose estimation \cite{courtois2024spiking}, there are still few spacecraft pose estimation datasets and solutions, and even fewer for event-based approaches.

SEENIC \cite{jawaid2023towards} is the first event-based dataset for spacecraft pose estimation, comprising both synthetic and real event data acquired on a robotic test bench. Although this dataset is suitable for simple deep learning direct approaches \cite{courtois2024spiking}, it remains challenging to use for complex or indirect machine learning approaches. The dataset has a limited variety of scenarios and a significant domain gap between the synthetically generated and real event data. SPADES \cite{rathinam2024spades} is another promising event-based dataset with mixed synthetic and real event data acquired on a robotic test bench, but it is still not available. Standard image datasets (grayscale, RGB, multispectral) released for pose estimation in the space industry mainly focus on reducing the domain gap between synthetic and real data. Some use a combination of synthetic and real images acquired on test benches, such as SPEED+ \cite{park2022speed+}, which contains 60,000 synthetic and 9,531 real images. Designed to train Formal Neural Networks (FNNs), most standard pose estimation datasets are composed of independent 2D images, unlike event-based data, which preserves temporal correlation. This loss of temporal information penalises their use for time-correlated event-based networks such as Spiking Neural Networks (SNNs).
As our main concern is operational cost behaviour rather than domain gap evaluation, and as no satisfactory event-based dataset is available, we decided to use the realistic RAPTOR dataset \cite{raptordataset}. With over 120,000 synthetic images, the RAPTOR dataset was built using a physics-based simulator aimed at achieving a high level of realism.

Recent advancements in neuromorphic hardware, such as SPLEAT \cite{abderrahmane2022spleat}, Pulsar \cite{pulsar2025}, Akida \cite{posey2022akida}, and SynSense Speck \cite{richter2023speck}, have shown promise for implementing Spiking Neural Networks (SNNs) in various applications, including space systems. However, these neuromorphic devices remain in the research phase and face significant challenges. Many existing neuromorphic platforms do not support a range of network architectures or have specific workflows that complicate deployment, due to hardware limitations in their design, processing capabilities, or availability. Additionally, some of these systems, such as SynSense Speck, operate asynchronously, which can complicate integration into real-time and critical applications such as satellite pose estimation.

While neuromorphic hardware has significant potential to improve computational efficiency and reduce power consumption, it is essential to design SNN architectures that align with these hardware processing paradigms to develop effective solutions.
For this purpose, we present the following contributions in this paper:
\begin{itemize}
    \item A comprehensive method to build and train an SNN architecture counterpart to a non-spiking neural network (FNN) architecture, targeting deployment on neuromorphic hardware.
    \item The first SNN solution for the RAPTOR dataset, targeting embedded keypoint prediction with heatmaps (see Fig. \ref{fig:Heatmapout}).
    \item A comprehensive computational and activity evaluation of our S-MoveNet-Lightning solution, refining the understanding of operational trade-offs in cost for Unet-like architectures between spiking and formal approaches.
\end{itemize}

%% file: sec2/2_related_work.tex
\section{Related Work}

\subsection{CNNs for Pose Estimation}
The computer vision community has shown increasing interest in efficient neural architectures for solving monocular pose estimation tasks in space-related applications. Research on CNNs can be divided into two main approaches: direct approaches \cite{wu20196d, park2024robust}, which estimate the pose of the target object using an end-to-end neural network, and indirect approaches \cite{geng2021bottom, wang2021gdr, MoveNet}, which rely on an intermediate representation that requires a pose refinement algorithm, generally based on Perspective-n-Point \cite{lepetit2009ep, terzakis2020consistently}. Some hybrid approaches, such as BPnP \cite{chen2020end}, explore the integration of Perspective-n-Point (PnP) pose computation into the learning process.

However, the stringent resource constraints of radiation-hardened platforms, such as limited computational resources and deployment chain restrictions, have led the community to prioritise computational efficiency.

Various optimisation techniques have been proposed to reduce the computational footprint of neural networks, making them more suitable for deployment on resource-constrained platforms. These methods include pruning \cite{lecun1989optimal}, quantisation \cite{wu2016quantized}, low-rank factorisation \cite{denton2014exploiting}, and knowledge distillation \cite{hinton2015distilling}, among others. The following section introduces an alternative approach to optimising the computational footprint of deep neural networks, which requires a different perspective.

\subsection{Spiking Neural Networks}
In a resource-constrained hardware environment, spiking neural networks (SNNs) offer a promising approach to energy-efficient, event-based processing. They process information only when relevant data occur, rather than at fixed frequencies, resulting in sparse data computation. This event-based paradigm can be applied not only to processing but also to information transmission and even sensors. Inspired by the biological functioning of neurons in the human brain, SNNs comprise neurons that communicate through discrete binary impulses over time, known as spikes. For computational purposes, time is discretised into \textit{timesteps}, during which spiking neurons accumulate input charge in their membrane potential, update their internal state, and emit spikes when a threshold is reached. With the Python library SpikingJelly \cite{fang2023spikingjelly}, spiking neurons can be used as an activation layer to construct SNNs. Various spiking neurons are available, but the most commonly used are Integrate and Fire (IF) neurons and Leaky Integrate and Fire (LIF) neurons, which introduce an additional leakage mechanism into the membrane potential over time. This can be set manually via a time constant or learned during training with Parametric Leaky Integrate and Fire (PLIF) neurons. This leakage enables improved temporal dynamics and more nuanced information processing.



\subsection{Event-based Processing}

To process information in SNNs, conventional data can be encoded using various methods, such as rate coding, time coding, and direct encoding. Although rate coding is widely used and can achieve state-of-the-art performance in many AI tasks, it generates a large number of spikes over many timesteps (often hundreds), which negatively affects computational cost, latency, and energy efficiency \cite{lemaire2022synaptic}. Time coding addresses the issue of timesteps by encoding information in the latency of spikes rather than their rate, resulting in fewer spikes overall. However, the temporal sparsity of these latency-encoded spikes can lead to longer processing times and can also be more difficult to implement \cite{kheradpisheh2020temporal}.

To overcome these limitations, direct encoding has been proposed. This method involves maintaining the dense and valued nature of the input to the first layer of the network, which uses a spiking neuron as its activation function (such as IF or LIF neurons). This approach enables the network to learn how to encode the data during training and to transform the dense and valued information into spikes, resulting in solutions with few timesteps and energy-efficient processing.

Moreover, SNNs face unique challenges in training and implementation. The non-differentiable nature of the thresholding mechanism, often computed using a Heaviside activation function, prevents the use of gradient descent-based training methods. Furthermore, the introduction of timesteps can cause latency issues and potential vanishing gradient problems, especially in large networks or those with many timesteps.

However, these challenges were addressed using specialised training approaches such as Surrogate Gradient Learning (SG) \cite{neftci2019surrogate} and Truncated Backpropagation Through Time (TBPTT) \cite{guo2023efficient}, which help to reduce training issues, particularly when using direct encoding techniques.

A key advantage of SNNs is their potential for energy efficiency. The event-driven nature of spike-based information flow leads to greater sparsity in activation outputs and fewer multiplication operations compared to traditional artificial neural networks, even when several timesteps are required to process the input data. If this sparsity is exploited during inference, the number of multiply-accumulate operations (MACs) can be reduced, with overall energy consumption primarily limited to memory access costs \cite{Lemaire2022}. However, spiking neurons incur higher memory access costs than conventional neural network activations due to the need to update the membrane potential, which limits potential energy gains. The energy saving ratio is therefore linked to the number of timesteps required for inference and the number of spikes relative to the number of neurons, reported as the activity.  
It has been shown that, when implemented on dedicated neuromorphic hardware, these networks can leverage their event-based nature and high sparsity to achieve approximately six times greater energy efficiency than conventional networks \cite{Dampfhoffer2022}\cite{Lemaire2022}.

\begin{figure*}
    \centering
    \includegraphics[page=2, clip, trim=0.9cm 3.5cm 1.2cm 8cm, width=1 \linewidth]{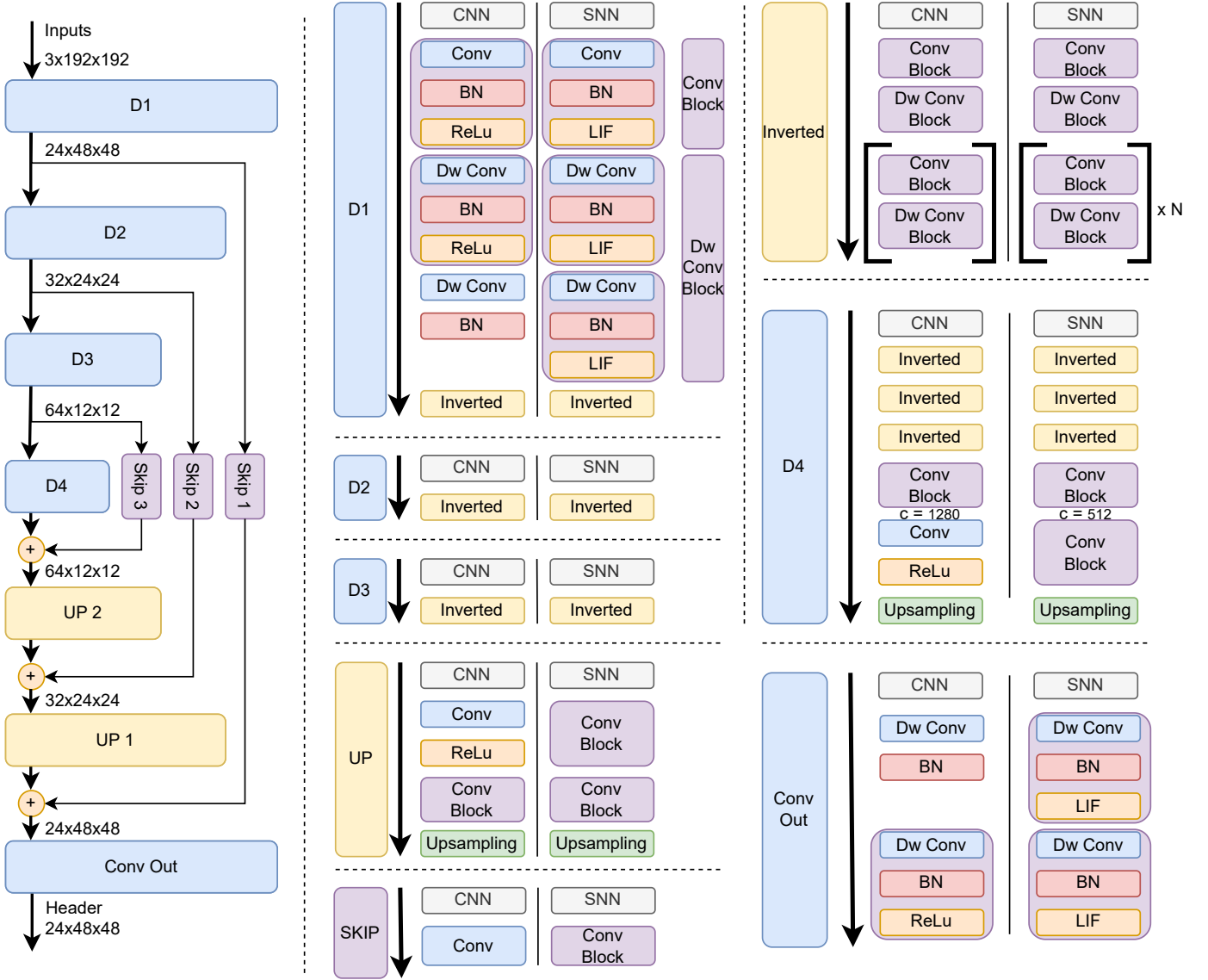}
    \caption{On the left, the complete architecture is depicted. On the right, the CNN and SNN implementations of each block are represented \cite{Jcourtois}.}
    \label{fig:MVNtransit}
\end{figure*}

\subsection{Neuromorphic hardware}
\label{spikeparadigm}

Neuromorphic hardware refers to computing systems designed to mimic the structure and function of the human brain, employing artificial neurons and synapses to process information, similar to SNNs.

Following this spiking paradigm, we define neuromorphic hardware as a device capable of sparse computation in an event-based manner, meaning it computes a neuron's output only when an input stimulus occurs. In our case, we base our reasoning on the limitation of a maximum of one spike per timestep per neuron.

With this paradigm, layers with spiking activation process incoming sparse feature maps composed of spikes and generate output sparse feature maps with spikes. It is important that the feature maps passed between each layer are sparse and consist of spikes. This processing method enables layers to handle binary sparse input, resulting in the sum of the weights associated with spikes rather than the weighted sum of the input, thereby reducing the number of multiplication operations (MAC) required. Additionally, the sparsity of the input allows the layer to process data in address event representation (AER) (the list of spikes from the previous layer) rather than the full feature maps, leading to computational savings and reduced energy consumption. Consequently, each layer or part of a network must be configured and adapted to avoid dense computation or manipulation of dense values. This paradigm provides guidelines to support the construction of neuromorphic hardware-friendly deployable networks.

Regarding this paradigm, the closest neuromorphic architecture is the SPLEAT accelerator \cite{abderrahmane2022spleat}, which supports up-sampling routing only in an experimental version, limiting us to an exploratory study in this paper. Designing an SNN with all these constraints in mind could reduce the complexity of deployment on neuromorphic hardware in future developments.

%% file: sec2/3_experiments.tex
\section{Experiments}
\label{sec:experiments}

In this section, we describe the implementation approach. Targeting an embedded deployable solution and evaluating a U-Net-like architecture with encoding and decoding components for pose estimation, we selected the MoveNet-Lightning architecture \cite{MoveNet}, available in the Vitis AI \cite{kathail2020xilinx} deployment tool Model Zoo. The SNN version of MoveNet was created using the SpikingJelly \cite{fang2023spikingjelly} libraries, which are compatible with PyTorch and enable the training of spiking neurons such as Parametric LIF. Parameter adjustments were made to ensure convergence during training for both the CNN and SNN approaches. The performances of the networks will be used for comparison purposes only, as the main objective is to evaluate the operational cost of this approach and refine the understanding of the factors that influence the computational cost ratio between SNN and FNN. This study provides insights into the impacts of transitioning to a spiking paradigm for pose estimation tasks in the context of space rendezvous, clarifying the path towards neuromorphic solutions.

Our experiments are based on the RAPTOR dataset \cite{raptordataset}, which consists of 120,000 synthetic samples specifically designed for Sentinel-3 satellite monocular pose estimation. The dataset provides fine statistical balance and realistic dynamic ranges using a physically based camera model. It comprises two subsets: nominal (D1) and perturbed (D2), with domain shifts in optical, sensor, scene, and target parameterisation. As robustness to the domain gap is outside the scope of our study, the benchmark is conducted on the D1 subset. For a complete space rendezvous, this study is also limited by the RAPTOR dataset to the approach phase (between 100 and 20 m) and a given 3D model (target satellite).

Our FNN and SNN approaches have been compared based on operational cost using the evaluation metrics in \cite{Lemaire2022}. These were compared to the \textit{torch summary} output and the Pose Score performance metrics defined in the Kelvin Satellite Pose Estimation Challenge (SPEC) 2021 (see section Competition Metric) \cite{kisantal2020satellite}.

\subsection{MoveNet Architecture}
\label{dataaug}
MoveNet \cite{MoveNet} is a Deep Learning architecture using heatmaps to accurately localize 2D keypoints for human pose estimation on the COCO dataset \cite{lin2014microsoft}. This architecture consists of two components:
\begin{itemize}
    \item A feature extractor: MobileNetV2 \cite{sandler2018mobilenetv2} with a Feature Pyramid Network (FPN) based on depthwise separable convolutions.
    \item Prediction heads: following CenterNet \cite{duan2019centernet} prediction scheme, this model predicts keypoint heatmaps, center heatmap, keypoint regression field and 2D per-keypoint offset field.
\end{itemize}
The model has been adapted to our use case (26 keypoints for the Sentinel-3 satellite versus 17 keypoints for humans originally). Architecture details are available in Figure \ref{fig:MVNtransit}. The pose performance shown in Table \ref{tab:model_metrics} was obtained through pose refinement using the SQPnP algorithm \cite{terzakis2020consistently}. 
The available version in the Vitis AI Model Zoo is called Lightning, which means the input spatial resolution is 192 × 192 pixels. The images in the RAPTOR dataset, originally at a native resolution of 2048 × 2048, are resized using OpenCV's AREA interpolation algorithm \cite{opencv_library} and normalised to a range between -1 and 1. To manage overfitting, we applied data augmentation techniques including blur, dropout, pixel jitter, contrast, brightness, hue, and saturation. Training was performed on an NVIDIA GeForce RTX 3090 using the PyTorch framework. The corresponding hyperparameters are listed in Table \ref{hyperparam}.

\subsection{S-MoveNet-Lightning Architecture}

The Spiking MoveNet architecture was built by focusing on the backbone part of the network, which has the greatest computational cost. The head will remain formal as the heatmap head has a negligible computational cost in this configuration and helps to achieve better regression performance. In this way, we target a deployment where the backbone is on a neuromorphic accelerator and the head on a CPU, as in this work \cite{courtois2024embedded} on automotive object detection.

To translate our backbone into the spiking paradigm, the ReLU activations were changed to PLIF neurons, which have a learnable decay factor. 
However, the MoveNet-Lightning architecture has many blocks which would break the spiking behavior of the network on neuromorphic hardware, as explained in section~\ref{spikeparadigm}.

To avoid this, the following change was applied to the architecture:
Spiking activations were added after each convolution layer that lacked an activation function, and batch normalisation layers were also included to increase the generalisation capability of the network.
As it is possible to fuse a convolution and batch normalisation layer into an equivalent convolution for inference, this enables the use of layer normalisation, for example, Convolution, batch normalisation, Spike neuron block, without breaking the spike paradigm \cite{Cordone2022}.
As depthwise convolution is supported by the SPLEAT accelerator \cite{abderrahmane2022spleat}, this is not an issue in our case but may be on other platforms.
The upsampling layers were set to 'nearest' mode to be equivalent to duplicating and redirecting output spikes towards the next layer, fitting the desired feature map spatial dimension without floating-point operations.
The spikes from the residual connection and the main path were merged with an addition layer. In this paradigm, this leads to training sparse feature maps of spikes with a maximum magnitude that can be greater than one. This can be processed by the neuromorphic hardware as a succession of spikes in temporal sequence.
In this way, upsampling layers and residual path fusion could also be used and interpreted as routing layers on neuromorphic hardware.
All changes are detailed in Figure \ref{fig:MVNtransit}.
 
Spiking neurons exhibit a temporal dependency in their membrane potential, which can be exploited using timesteps. For static inputs, timesteps can be unrolled over the same data to enhance network performance, at the cost of increased latency and computational demand \cite{castagnetti2023spiden}. Direct encoding achieves better performance with fewer timesteps than rate coding, so the network is configured for 5 timesteps over the same input images. As the backbone is the only spiking component of the network, we unrolled timesteps over it and provided the head with the average output across the 5 timesteps.

After training, the mean activity of the network – defined as the average number of spikes per timestep divided by the number of neurons – determines the average number of multiply-accumulate operations (MACs) during inference. A key challenge in training a Spiking Neural Network (SNN) is to ensure sufficient activity for network convergence while avoiding excessive MACs. To regulate this activity level, we added a term to the loss function to account for the targeted activity, denoted as \(act\_loss\).

\begin{equation} 
\label{eq:21} 
    act\_loss = (spike_{nb} - net_{dim} * \frac{act_{tgt}}{100})^2 * 10^{-9} 
\end{equation} 
In this equation, \(spike_{nb}\) refers to the total number of processing spikes per timestep, \(net_{dim}\) denotes the number of neurons in the network, and the \(act_{tgt}\) hyperparameter represents the targeted overall activity level. The factor \(10^{-9}\) is used to weight this squared error relative to other terms in the loss function. The \(act_{tgt}\) was arbitrarily set to 30\%. 
Training of the S-MoveNet-Lightning model was performed on an NVIDIA RTX A6000 using the PyTorch and SpikingJelly frameworks. 
The hyperparameters for this training are detailed in Table \ref{hyperparam}. The data augmentation techniques described in the previous section \ref{dataaug} were also employed.

\begin{table}[ht]
    \centering
    \vspace{0.3cm}
    \begin{tabular}{|l|c|c|}
    \hline
    \textbf{Parameter} & \textbf{Lightning} & \textbf{SNN} \\
    \hline
    width\_mult        & 1.00 & 1.00     \\
    img\_size          & 192 & 192       \\
    feature\_map\_size & 48 & 48         \\
    weight\_decay      & 5.e-4 & 0       \\
    learning\_rate     & 1e-3 & 5e-4     \\
    batch\_size        & 256 & 100      \\
    epochs             & 120 & 200      \\
    timesteps          & - & 5          \\
    \hline
    \end{tabular}
    \vspace{0.4cm}   
    \caption{Hyperparameters of different MoveNet versions on the RAPTOR dataset.}
    \label{hyperparam}
\end{table}

\subsection{Quantization}

When targeting embedded deployment, quantisation is a key step to reduce model size and computational cost. It involves converting model weights and activations from floating-point precision (FP32) to lower precision formats, such as INT8. This process can significantly reduce memory usage and increase inference speed, especially on hardware with limited resources, but generally at the cost of accuracy.
Post-Training Quantisation (PTQ) uses a calibration dataset to compute the scale and offset, enabling determination of a low-precision encoding (8-bit in this case). This method is fast and simple but may result in a significant loss of accuracy.
For information, we present the results of INT8 PTQ applied to the MoveNet-Lightning and S-MoveNet-Lightning architectures. We performed a naive conversion from FP32 to INT8, using 4 bits for the integer part and 4 bits for the fractional part. All weights, biases, membrane potentials, thresholds, decay, and activation values were quantised to INT8. The results are shown in Table \ref{tab:model_metrics}.

\subsection{Comparison Tool}

To compare the computational cost of S-MoveNet-Lightning with MoveNet-Lightning, we used the \textit{torch summary} tool. This tool provides a detailed summary of the model, including the number of parameters, the number of operations (MACs), and the memory footprint. The MACs count is particularly useful for understanding the computational complexity of the model, as it indicates how many multiply-accumulate operations are required for a single forward pass through the network. However, the summary does not show the difference or proportion between Multiply and Add operations, which is important for a more precise understanding of the network's computational cost.
For this reason, we combined the information from \textit{torch summary} with the operational cost evaluation equation in~\cite{Lemaire2022}. This work provides metrics to compare FNN and SNN in terms of synaptic operation, addressing, and memory cost to evaluate energy consumption. As we focus here on computational cost and convolutional networks, we use only the MACC and ACC operations counts to compare the two approaches. Moreover, the paper does not consider depthwise convolutions, which are used in the MoveNet-Lightning architecture. To address this, we have adapted the equation for convolution when a depthwise convolution occurs in our network, as follows:

\begin{equation}
    \label{eq-ACC_MAC_SNN_FNN}
    \begin{split}
    \text{MAC}_\text{DW\_Conv-l}^\text{FNN} & = C_\text{out} \times H_\text{out} \times W_\text{out} \times H_\text{kernel} \times W_\text{kernel}\\
    \text{ACC}_\text{DW\_Conv-l}^\text{FNN} & = C_\text{out} \times H_\text{out} \times W_\text{out} \\
    \text{MAC}_\text{DW\_Conv-l}^\text{SNN} & = T \times C_\text{out} \times H_\text{out} \times W_\text{out} \\
    \text{ACC}_\text{DW\_Conv-l}^\text{SNN} & = \theta_\text{l - 1} \times (\lceil \frac{H_\text{kernel}}{S} \rceil)\times (\lceil \frac{W_\text{kernel}}{S} \rceil) \\ & + T \times C_\text{out} \times H_\text{out} \times W_\text{out} + \theta_\text{l} \\
    \end{split}
\end{equation}

where \(C_\text{out}\) is the number of output channels, \(H_\text{out}\) and \(W_\text{out}\) are the height and width of the output feature map, \(H_\text{kernel}\) and \(W_\text{kernel}\) are the height and width of the convolution kernel, \(S\) is the stride, \(\theta_\text{l - 1}\) is the number of incoming spikes, \(\theta_\text{l}\) is the number of output spikes, which is equal to the number of reset operations, and \(T\) is the number of timesteps.

In a network without residual or skip connections, the generated and processed spikes are the same. 
As this network includes upscaling phases and addition layers, the computation of network activity must focus on the number of spikes processed by each layer rather than the number generated. As noted previously, concatenating these paths can result in a sum of spikes, producing spike feature maps with values clamped between 0 and N (2 in our case). Even if neuromorphic hardware could interpret this as N consecutive spikes, the additional cost would be obscured if generated spikes are used to compute network activity. In this paper, network activity is defined as the sum of the processed spikes across the entire network, divided by the total network dimensions.

\label{sec:results}
\section{Results}
\begin{table}[b]
    \centering
    \begin{tabular}{|l|l|l|l|l|l|l|}
    \hline
    \textbf{Model} & \textbf{Input} & \textbf{Arithmetic} & \textbf{MACs} &  \textbf{Pose score $\downarrow$} \\
    \hline
    SPNv2-B0 \cite{park2024robust, raptordataset}   & 512x512 & FP32        & 3.7G & \textbf{0.6} \\
    \hline
    MoveNet-Lightning             & \textbf{192x192} & FP32        & 239M & 1.04 \\
    \hline
    MoveNet-Lightning             & \textbf{192x192} & \textbf{INT8}  & 239M & 3.16 \\
    \hline
    S-MoveNet-Lightning             & \textbf{192x192} & FP32        & \textbf{108M} & 2.34 \\ 
    \hline
    S-MoveNet-Lightning             & \textbf{192x192} & \textbf{INT8}   & \textbf{108M} & 3.19 \\ 
    \hline
    \end{tabular}
    \caption{Performance metrics (translation, rotation and pose) of evaluated models on RAPTOR testing subset (D1).}
    \label{tab:model_metrics}
\end{table}

The results of the network quantised versions on the D1 subset (nominal cases) of the RAPTOR dataset provide insights into the impact of quantisation (from 32-bit floating point to 8-bit integer). Maintaining on-board floating-point performance is challenging for this type of network; both networks experience a significant drop in performance, highlighting the need for more advanced quantisation techniques such as Quantisation-Aware Training (QAT).

\begin{figure*}[ht]
    \centering
    \includegraphics[width=1 \linewidth]{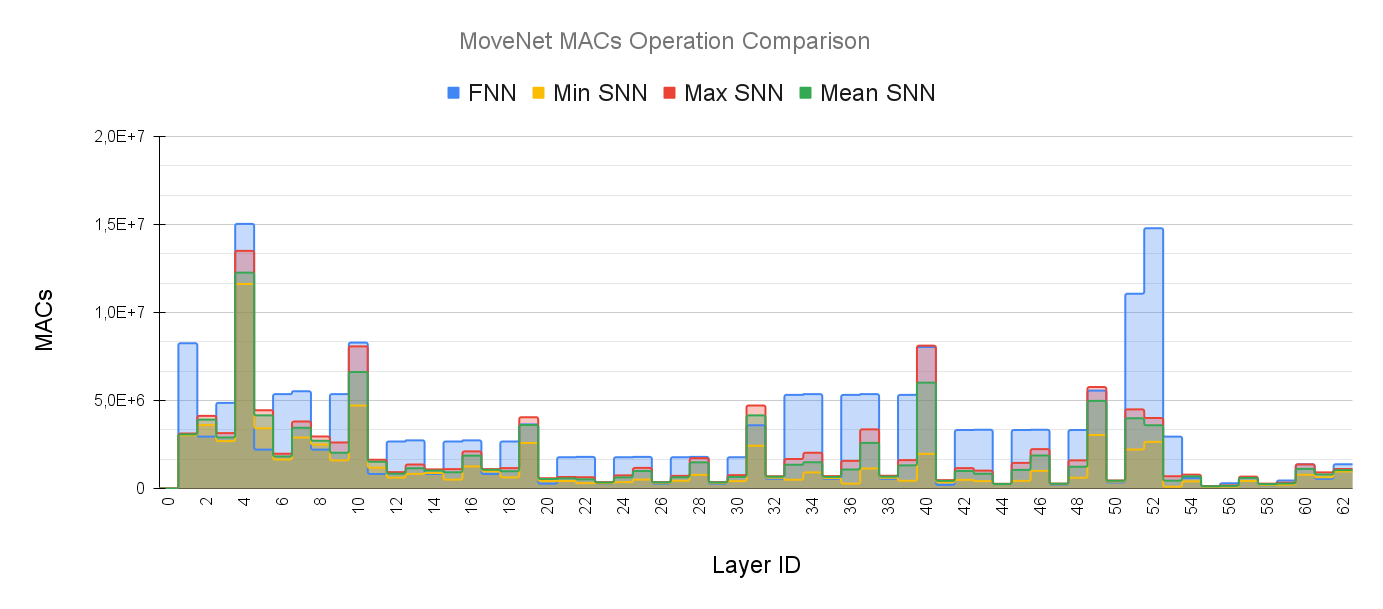}
    \caption{MACs operation detailed for each layer. The S-MoveNet-Lightning has 3 representations covering the mean, min and max of the network activity over the D1 testing subset.}
    \label{fig:MVNlayercomp}
\end{figure*}

Regardless of the quantisation evaluation, the following results emerge from the study:
\begin{itemize}
    \item The original implementation available in AMD's Model Zoo (MoveNet-Lightning) shows a significant drop in pose estimation performance on the RAPTOR dataset, in both Formal and Spiking versions, compared to the baseline network SPNv2-B0~\cite{park2024robust}. This was expected, as we have reduced the input dimension and used lighter networks, resulting in 3.461G and 3.59G fewer MACs for the FNN and SNN respectively, as shown in Table \ref{tab:model_metrics}.
    \item Compared to the FP32 MoveNet-Lightning version, our S-MoveNet-Lightning requires 2.2 times fewer MACs, including the 5 timesteps.
    \item A detailed comparison of the synaptic operation cost between the two networks, layer by layer, is available in Fig \ref{fig:MVNlayercomp}.
\end{itemize}

Figure \ref{fig:Heatmapout} presents sample images from the RAPTOR dataset alongside the raw stacked heatmaps predicted by the proposed SNN approach, illustrating the method's performance. The top two images from the training set show complex scenes with the Earth in the background and squares added by dropout data augmentation. Notably, the network is able to focus on the satellite, even when it overlaps with the Earth, and ignores the squares, demonstrating its generalisation capabilities.

The approach studied here shows that it is possible to estimate the pose of a satellite using a spiking neural network with a remarkable compression rate. However, an analysis of event-driven input sequences is still required to confirm the operational performance of this type of neural network.

The MoveNet-Lightning approach, in its current form, does not achieve the performance levels required for this space rendezvous mission. It is important to note that the average error for distance ranges between 20 and 100 metres is significantly affected by errors at longer distances, which are less critical. The results presented in Table 1 are intended solely for relative comparison. This study aims to provide a foundation for evaluating the impact of such an adaptation on a lightweight pose estimation architecture. Future work could extend this analysis to more complex architectures, such as SPNv2 \cite{park2024robust}, to achieve performance levels closer to the desired standards.

Using the \textit{torch summary} tool to output the multiplication/addition operations count of the network, our S-MoveNet-Lightning network shows a total of 995.03 million mult-add operations for one inference. However, this result includes the backbone pass-through repeated over the five timesteps, the head, and assumes the network is computed in a dense manner. This figure should be balanced with the mean activity of the network after training to better estimate how many mult-add operations will be processed on neuromorphic hardware, given their sparse computation capability. After training, the mean activity in the spike features of the S-MoveNet-Lightning network is 26.68\%. This activity reflects the number of spikes to be processed on the neuromorphic hardware, providing an estimate of the computational operations that must be considered to balance the total MAC operations reported by \textit{torch summary}. Accordingly, only a mean of 108 million MAC operations will be executed on dedicated hardware.

%% file: sec2/4_conclusion.tex
\section{Discussion}
\label{sec:discussion}

The post-processing described in the original MoveNet paper (using the other heads) has not been integrated or validated, and may be implemented in the future. We chose to limit the study to a single measurement point using raw heatmaps to ensure the quality of our benchmark. This indicates that there is still potential for performance improvement in this area.

Regarding S-MoveNet-Lightning, several critical factors were addressed in this study to achieve the reported results. Without sufficient data augmentation, both SNN and CNN versions of the MoveNet architecture exhibited significant overfitting. The proposed data augmentation strategy effectively regularises the model, leading to smoother performance convergence. Additionally, the SNN approach requires significantly longer training times due to its timestep-based processing.
Moreover, if the dataset consisted of video samples, unrolling the five timesteps over five consecutive images, the temporal correlation between images could be exploited. This would make the network equivalent to a one-timestep network, thereby reducing the overall cost of MAC operations.

On neuromorphic hardware, only spikes generate MAC operations in the feature maps for synaptic operation. However, this activity measurement is based on test set inference statistics. Specifically, the minimum and maximum activity boundaries for S-MoveNet-Lightning are set between 16.46\% and 30.90\%, leading to a range of 78M to 125M MAC operations on neuromorphic hardware. These boundaries contrast with the MAC operations of counterpart CNN networks. S-MoveNet-Lightning provides a MAC reduction ranging from 3.1 to 1.9 times during inference for the floating-point version.
Rather than focusing solely on the minimum and maximum activity statistics, we decided to evaluate the impulse response of the network. Feeding the network with input frames where each pixel receives the maximum value at each timestep shows a 1.3-fold mean activity increase reaching 35.59\%. This leads to a 'worst-case' scenario of 141M MACs, which is still considerably lower than the 239M MACs of the formal version.
The minimum, maximum, and impulse tests are easy to implement and not often used in the literature, but they provide a more comprehensive understanding of the spiking behaviour of the network, which is even more important in critical environments such as space applications.
Furthermore, the \textit{torch summary} tool does not differentiate between multiplication and addition operations, presenting only their total sum. However, thanks to the synaptic operation metrics, we could extract this information.
The reduction of synaptic operations from 239M to 108M involves 211M fewer MACs but 81M more ACCs on average. The SNN has 22M MACs and 86M ACCs.
This study focused solely on MACs, but as SNNs incur additional memory costs due to membrane potential updates, energy estimation cannot be extrapolated solely from MAC operations, as memory energy cost is significant \cite{Lemaire2022}. The energy estimation of S-MoveNet-Lightning will be addressed in future research.
\section{Conclusion}

This work has presented a comprehensive evaluation of a spiking implementation of the MoveNet-Lightning architecture in the context of space applications. Through our investigation of S-MoveNet-Lightning on the RAPTOR dataset, we have demonstrated the potential of spiking neural networks as an alternative to traditional formal neural networks for efficient embedded satellite applications.
Our experimental results show that the proposed S-MoveNet-Lightning architecture requires approximately 2.2 times fewer multiplication and accumulation operations than its formal counterpart, reducing computational demands from 239M to 108M MACs. The analysis of network activity, ranging from 16.46\% and 30.90\%, provides valuable insights into the operational behaviour of spiking neural networks in U-Net-like architectures. Even in worst-case scenarios with maximum input stimulation, the activity remains bounded at 35.59\%, resulting in no more than 141M MACs, which is still considerably lower than the formal version. This demonstrates the robustness of the approach in maintaining computational efficiency across varying operational conditions.
Several challenges remain to be addressed in future work. First, more advanced quantisation techniques, such as Quantisation-Aware Training (QAT), should be explored to mitigate the significant performance degradation observed with post-training quantisation. Second, evaluation with event-based datasets would provide more comprehensive insights into the temporal processing capabilities of SNNs. Finally, actual deployment on neuromorphic hardware would enable accurate energy consumption measurements beyond the MAC-based estimations presented in this study.